\documentclass[11pt,a4paper]{article}
\usepackage[hyperref]{emnlp-ijcnlp-2019}
\usepackage{times}
\usepackage{booktabs}
\usepackage{latexsym}
\usepackage{graphicx}
\usepackage{url}
\usepackage{multirow}
\usepackage{array}
\usepackage{tabularx}
\usepackage{caption,tabularx,booktabs}

\aclfinalcopy 

\title{\emph{Y'all should read this!}\\
Identifying Plurality in Second-Person Personal Pronouns in English Texts}

\author{Gabriel Stanovsky \\
  University of Washington \\
  Allen Institute for Artificial Intelligence \\
  {\tt gabis@allenai.org} \\\And
  Ronen Tamari\thanks{~ Work done during an internship at the Allen Institute for Artificial Intelligence.} \\
  Hebrew University of Jerusalem \\
  {\tt ronent@cs.huji.ac.il} \\}

\date{}

\begin{document}
\maketitle


\begin{abstract}
  Distinguishing between singular and plural ``you'' in English is a challenging task
  which has potential for downstream applications, such as machine translation or coreference resolution.
  While formal written English does not distinguish between these cases, other languages (such as Spanish), as well as other dialects of English (via phrases such as ``y'all"), do make this distinction.
  We make use of this to obtain distantly-supervised labels for the task on a large-scale in two domains.
  Following, we train a model to distinguish between the single/plural `you', finding that although in-domain training achieves reasonable accuracy ($\geq$ 77\%), there is still a lot of room for improvement, especially in the domain-transfer scenario, which proves extremely challenging.
  Our code and data are publicly available.\footnote{\url{https://github.com/gabrielStanovsky/yall}}
\end{abstract}

\section{Introduction}

The second-person personal pronoun (e.g., ``you'' in English) is used by a speaker when referring to active participants in a dialog or an event.
Various languages, such as Spanish, Hebrew, or French, have different words to distinguish between singular ``you'' (referring to a single participant) and plural ``you'' (for multiple participants).
Traditionally, English has made this distinction as well. The now archaic ``thou'' indicated singular second-person, while ``you'' was reserved for plural uses.
The last several hundred years, however, have seen modern formal written English largely abandoning this distinction, conflating both meanings into an ambiguous ``all-purpose you''~\cite{maynor2000battle}.

In this work, we are interested in the following research question: \emph{How can we automatically disambiguate between singular and plural ``you"?}

\begin{figure}
  \includegraphics[width=\linewidth]{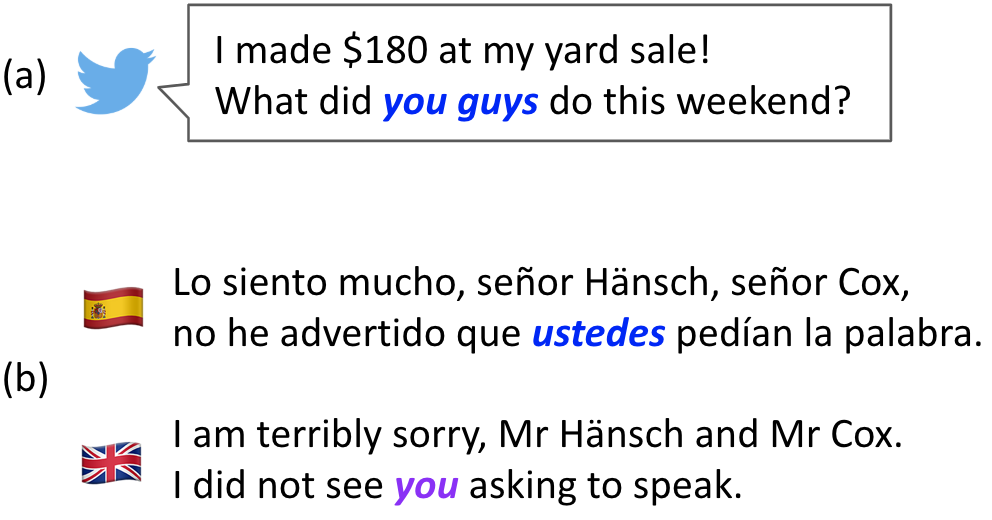}
  \caption{We use two sources for distant-supervision for singular (marked in purple) versus plural (marked in blue) second person pronouns:
  (a) we find colloquial uses on Twitter, and (b) through alignment with Spanish, which formally
  distinguishes between the cases.}
  \label{fig:sources}
  \end{figure}

\begin{figure*}[tb!]
  \center
  \includegraphics[width=0.8\linewidth]{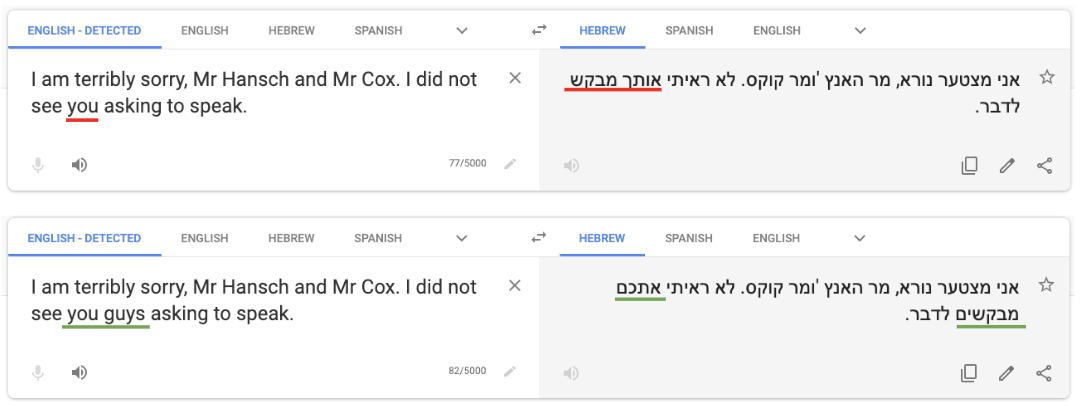}
  \caption{An example translation from English to Hebrew (Google Translate, Aug. 21, 2019).
    The first sentence depicts
    wrong interpretation of ``you'' resulting in a non-grammatical Hebrew translation, due to wrong inflections of pronoun and verb (marked in red).
    Both issues are fixed when changing ``you'' to ``you guys'' in English in the
    second example (marked in green).
  }
  \label{fig:mt}
  \end{figure*}


While this topic has received much attention in linguistic literature \cite{jochnowitz1983another,Tillery1998YallIO,maynor2000battle,haspelmath2013politeness,molina2016}, it has been largely unexplored in the context of computational linguistics, despite its potential benefits for downstream natural language processing (NLP) tasks.
For example, distinguishing between singular and plural ``you'' can serve as additional signal when translating between English and a language which formally makes this
distinction. See Figure \ref{fig:mt} where an error in interpreting a plural ``you'' in the source English text results in a non-grammatical Hebrew translation.
This example can be amended by replacing ``you'' with the informal ``you guys''. 


\begin{table*}[!t]
  \centering
  \small
\begin{tabular}{p{2cm}p{10cm}p{2cm}}
  \toprule
  \textbf{Domain} & \textbf{Example} & \textbf{Plurality} \\
  \midrule
  Twitter & \# goodnight \#twittersphere $<$$3$ I love \textbf{y'all}! Including @{anonimized}. Even if she hates me. $<$$3$ & \emph{Plural} \\
  (masked)  &  \# goodnight \#twittersphere $<$$3$ I love \textbf{you}! Including @{anonimized}. Even if she hates me. $<$$3$ & \\
  \midrule
  Twitter & ! @{anonimized}, Happy anniversary of entering the world!  Look how much \textbf{you} have done!& \emph{Singular} \\
  \midrule
  Europarl & I am terribly sorry, Mr Hansch and Mr Cox. I did not see \textbf{you} asking to speak. & \emph{Plural} \\
  \midrule
  Europarl & I should be very grateful, Mrs Schroedter, if \textbf{you} would actually include this proposed amendment in the part relating to subsidiarity in your positive deliberations. & \emph{Singular} \\
  \bottomrule
\end{tabular}
\caption{Examples from our two domains. Twitter is  informal, includes hashtags, mentions (anonymized here), and plural ``you'' (e.g., ``y'all'' in the first example), which we mask
  as a generic ``you'' as shown in the second row. In contrast, Europarl is formal and ``you'' is used for plural (third example), as well as singular uses (last example).}
\label{tab:examples}
\end{table*}

To tackle this task, we create two large-scale corpora annotated with distantly-supervised binary labels distinguishing between singular and plural ``you'' in two different domains (see Figure \ref{fig:sources}).
First, we regard Twitter as a noisy corpus for informal English and automatically identify speakers who make use of
an informal form of the English plural ``you'', such as ``y'all" or ``you guys", which are common in American English speaking communities \cite{katz2016speaking}. We record a \emph{plurality} binary label, and mask the tweet by replacing these with the generic ``you''.
Second, we use the Europarl English-Spanish parallel corpus \cite{Koehn2005EuroparlAP}, and identify cases where the formal plural Spanish second-person pronoun aligns with ``you'' in the English text.

Finally, we fine-tune BERT \cite{bert} on each of these corpora. We find that contextual cues indeed allow our model to recover the correct intended use in
more than 77\% of the instances, when tested in-domain. Out-of-domain performance drops significantly, doing only slightly better than a random coin toss. This could indicate that models are learning surface cues which are highly domain-dependent and do not generalize well.

Future work can make use of our corpus and techniques to collect more data for this task, as well as incorporating similar predictors in downstream tasks.


\section{Task Definition}
Given the word ``you'' marked in an input text, the task of \emph{plurality identification}
is to make a binary decision whether this utterance refers to:
\begin{itemize}
\item A single entity, such as the examples in rows 2 or 4 in Table \ref{tab:examples}.
\item Multiple participants, such as those
  referred to in the third line in Table
  \ref{tab:examples}.
\end{itemize}

In the following sections we collect data for this task and develop models for its automatic prediction.

\section{Distant Supervision: The \emph{y'all} Corpus}
\begin{table}[!tb]
  \centering
  \begin{tabular}{@{}lrr@{}}
    \toprule
    \textbf{} & \multicolumn{1}{l}{\textbf{Twitter}} & \multicolumn{1}{l}{\textbf{Europarl}} \\ \midrule
    Train & 58963 & 11249 \\
    Dev & 7370 & 1405 \\
    Test & 7370 & 1405 \\
    \textbf{Total} & \textbf{73703} & \textbf{14059} \\ \bottomrule
  \end{tabular}
  \caption{Number of instances in our two corpora. Each of the partitions (train, dev, test) is equally split between plural and singular second-person personal pronouns.}
  \label{tab:corpora}
  \end{table}

Manually collecting data for this task on a large-scale is an expensive and involved process.
Instead, we employ different techniques to obtain distantly supervised labels in two domains, as elaborated below.
These are then randomly split between train (80\% of the data), development, and test (10\% each).
See Table \ref{tab:corpora} for details about each of these datasets, which we make publicly available.

\subsection{The Twitter Domain}
As mentioned in the Introduction, English speaking communities tend to maintain the singular versus plural ``you'' distinction by introducing idiosyncratic phrases which
specifically indicate a plural pronoun, while reserving ``you'' for the singular use-case.
We operationalize this observation on a large Twitter corpus \cite{Cheng2010YouAW} in the following manner:
\begin{itemize}
\item
  First, we identify speakers who use an informal plural ``you'' at least once in any of their tweets.\footnote{We use a fixed list of informal plural ``you'', including \emph{you guys}, \emph{y'all} and \emph{youse}. See \url{https://en.wikipedia.org/wiki/You\#Informal_plural_forms} for the complete list.}
\item
  Following, we assume that these users speak an English dialect which distinguishes between singular and plural second-person pronouns, interpreting
  their ``you'' as a singular pronoun. See the first two tweets in Table \ref{tab:examples}, for an example of these two uses by the same user.
\item
  Finally, we mask out the plural pronoun in each of their tweets by replacing it with a generic ``you'' (see the second row in Table \ref{tab:examples}). This allows us to test whether models
  can subsequently rely on contextual cues to recover the original intention.
\end{itemize}

This process yields about 36K plural instances, which we augment with 36K singular instances from the same users, to obtain a corpus which is balanced between the two classes.

\begin{figure}[tb!]
  \includegraphics[width=\linewidth]{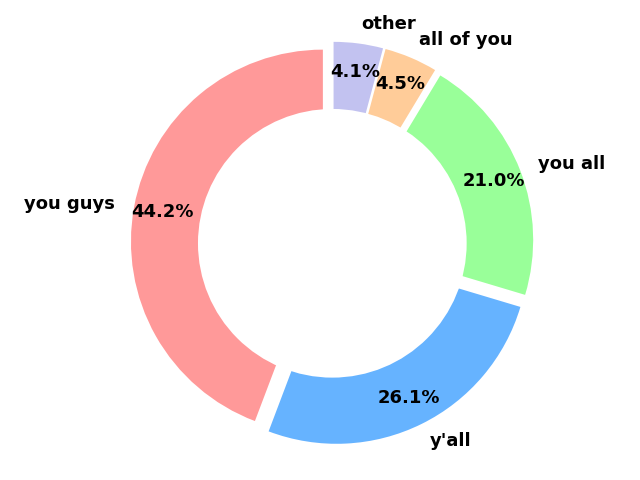}
  \caption{Histogram distribution of informal plural ``you'' forms in the development partition of our Twitter corpus.}
  \label{fig:hist}
\end{figure}

\begin{figure}[tb!]
  \includegraphics[width=\linewidth]{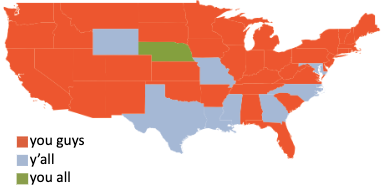}
  \caption{Variation in the most common phrase used for plural ``you'' in our Twitter corpus across states in the continental United States.}
  \label{fig:us-plot}
  \end{figure}

\paragraph{Data analysis}
Our Twitter corpus was composed of U.S. based users, and included geo-locations for 36.8K of the  plural tweets. This allows for several interesting observations.
First, Figure \ref{fig:hist} shows the distribution of informal plural ``you'' phrases in our corpus (before masking).
Second, using the tweets geo-location, we can plot the geographical variation in usage.
Figure \ref{fig:us-plot} shows the most common term for plural ``you'' in each state in the continental United States.
While most of the U.S. prefers ``you guys'', southern states (such as Texas or Louisiana) prefer ``y'all''. \newcite{katz2016speaking} reached similar conclusions through a large-scale online survey. Interestingly, our survey of Twitter usage differs from theirs for several Midwestern states, such as Wyoming or Nebraska.

\paragraph{Quality estimation.}
We evaluate the assumption we made above (i.e., that users reserve ``you'' for the singular case)
by asking an English native-speaker to annotate a sample of 100 singular ``you'' instances from our Twitter corpus.
In 70\% of the instances, the annotator agreed that these indeed represent a singular ``you'', leaving the other 30\% to either ambiguous or inconsistent usage
(i.e., sometimes ``you'' \emph{is} used for a plural use-case).
Overall, while there is a non-negligible amount of noise in the Twitter domain, in Section \ref{sec:evaluation} we show that the signal is strong enough for models
to pick up on and achieve good accuracy.

\subsection{The Europarl Domain}
Another method to obtain distant supervision for this task is through an alignment with a language which distinguishes between the two usages of the pronoun.
To that end, we use the Spanish and English parallel texts available as part of Europarl~\cite{Koehn2005EuroparlAP}, a large corpus containing aligned sentences from meeting transcripts of the
European parliament.

As these texts originate from a formal setting, we expect to find much less colloquial phrases. Indeed, the term ``y'all'' does not appear at all, while ``you guys'' appears only
twice in about 2 million sentences.
Instead, we rely on the gold alignment with Spanish sentences, which have a formal plural ``you'' - \emph{ustedes} or \emph{vosotros}.
We find Spanish sentences which have exactly one plural ``you'' and which aligns with an English sentence containing exactly one ``you''.
This process yields about 7K sentences which we mark with a ``plural'' label. Similarly to the Twitter domain, we augment these with the same amount of singular ``you'', found in the same manner; by tracing a Spanish singular ``you'' to a single English ``you''. This process yields a balanced binary corpus.

\paragraph{Quality estimation}
We sampled 100 instances from the Europarl domain to estimate the quality of our binary labels.
Unlike the Twitter domain, in Europarl we rely on gold alignments and cleaner text. As a result, we found that about 90\% of the labels agree with a human annotator, while the remaining 10\% were ambiguous.

\subsection{Discussion}
The distinction between plural and singular ``you" in English is an instance of a more general phenomenon. Namely, semantic features are expressed in varying degrees of lexical or grammatical explicitness across different languages.

For instance, languages vary in grammatical tense-marking \cite{wolfram1985variability}, from languages with no morphological tense, such as Mandarin  \cite{wang2015oxford}, to languages with 9 different tense-marking morphological  inflections~\cite{comrie1985tense}.
Similarly, languages vary in gender-marking in pronouns, from gender-less Turkish, to languages with six genders or more~\cite{awde1997chechen}.

The two data collection methods we presented here, finding colloquial explicit utterances on social media, and alignment with another language, may also be applicable to some of these phenomena and present an interesting avenue for future work.


\section{Model}
We fine-tune the BERT-large pretrained embeddings \cite{bert}\footnote{Using Hugging Face's implementation: \url{https://github.com/huggingface/pytorch-transformers}} on the training partition of each of our domains (Twitter and Europarl), as well as on a concatenation of both domains (Joint). We then classify based on the [CLS] token in each of these instances.
We use a fixed learning rate of $2e-5$ and a batch size of 24. Training for 10 Epochs on a Titan X GPU took about 3 hours for the Twitter domain, 2 hours for the Europarl domain
and roughly 5 hours for the Joint model.

\section{Evaluation}
\label{sec:evaluation}
We test models trained both in and out of domain for both parts of our dataset (Twitter and Europarl) as well as a joint model, trained on both parts of the dataset.
We use accuracy (percent of correct predictions), as our dataset is binary and both classes are symmetric and evenly distributed.
Our main findings are shown in Table \ref{tab:eval}.
Following, several observations can be made.

\begin{table}[]
  \centering
  \begin{tabular}{@{}lrr@{}}
    \toprule
    \textbf{test $\rightarrow$} & \multicolumn{1}{l}{\multirow{2}{*}{\textbf{Europarl}}} & \multicolumn{1}{l}{\multirow{2}{*}{\textbf{Twitter}}} \\
    \textbf{train $\downarrow$} & \multicolumn{1}{l}{} & \multicolumn{1}{l}{} \\ \midrule
    Europarl & 77.1 & 56.8 \\
    Twitter & 56.3 & \textbf{83.1} \\
    Joint & \textbf{77.5} & 82.8 \\ \bottomrule
  \end{tabular}
  \caption{Accuracy (percent of correct predictions)
    of our fine-tuned BERT model, tested both in- and out-of- domain.
    Rows indicate train corpus, columns indicate test corpus.
    Bold numbers indicate best performance on test corpus.
  }
  \label{tab:eval}
  \end{table}ß

\paragraph{In-domain performance does significantly better than chance.} For both domains, BERT achieves more than 77\% accuracy. Indicating that
the contextual cues in both domains are meaningful enough to capture correlations with plural and singular uses.

\paragraph{Out-of-domain performance is significantly degraded.} We see significant drop in performance when testing either model on the other part of the dataset. Both models
do only slightly better than chance. This may indicate that the cues for plurality are vastly different between the two domains, probably due to differences in vocabulary, tone, or
formality.

\paragraph{Training jointly on the two domains maintains good performance, but does not improve upon it.} A model trained on both the Twitter and Europarl domains
achieves the in-domain performance of each of the individual in-domain models, but does not improve over them. This may indicate that while BERT is expressive enough to model both domains, it only picks up on surface cues in each and does not generalize across domains. As a result, robustness is questionable for out-of-domain instances.

\section{Related Work}
Several previous works have touched on related topics.
A few works developed models for understanding the second-person pronoun within
coreference resolution frameworks \cite{purver2009cascaded,zhou2018they}.
Perhaps most related to our work is  \newcite{gupta2007disambiguating}, who have tackled the orthogonal problem of  disambiguation between generic (or editorial) ``you'' and referential ``you''.

To the best of our knowledge, we are the first to deal with plurality identification in second-person personal pronouns in English.

\section{Conclusion and Future Work}
We presented the first corpus for the identification of plurality in second-person personal pronouns in English texts.
Labels were collected on a large scale from two domains (Twitter and Europarl) using different distant-supervision techniques.

Following, a BERT model was fine-tuned on the labeled data, showing that while  models achieve reasonable in-domain performance, they significantly suffer from domain transfer,
degrading performance close to random chance.
Interesting avenues for future work may be to extend this data to new domains, develop more complex models for the task (which may achieve better cross-domain performance), and
integrating plurality models in downstream tasks, such as machine translation or coreference resolution.

\section*{Acknowledgements}
We thank the anonymous reviewers for their many helpful comments and suggestions.

\bibliography{emnlp-ijcnlp-2019}
\bibliographystyle{acl_natbib}

\appendix

\end{document}